\documentclass[10pt,twocolumn,letterpaper]{article}

\usepackage[pagenumbers]{cvpr} 

\usepackage[dvipsnames]{xcolor}

\newlength\savewidth

\makeatletter\renewcommand\paragraph{\@startsection{paragraph}{4}{\z@}
	{.1em \@plus1ex \@minus.2ex}{-.5em}{\normalfont\normalsize\bfseries}}\makeatother

\definecolor{mygray}{gray}{.9}
\definecolor{light-gray}{gray}{0.72}

\usepackage{booktabs}       %
\usepackage{amsmath,amsfonts,amsthm}   %
\usepackage{mathtools} 
\usepackage[font=footnotesize]{caption}
\usepackage{caption}
\usepackage{algorithm}
\usepackage{algorithmic}
\usepackage{wrapfig,lipsum,booktabs}
\usepackage{graphics}
\usepackage{enumitem}
\usepackage{colortbl}
\usepackage{xcolor}
\definecolor{mygray}{gray}{.9}
\usepackage{booktabs}
\usepackage{xspace}
\usepackage{bbding}
\usepackage{tikz}
\usepackage{comment}
\usepackage{enumitem}
\usepackage{amsmath,amssymb}
\setlist{nolistsep}
\newtheorem*{theorem*}{Theorem}

\usepackage{amsmath,amsfonts,bm}

\def\eqref#1{equation~\ref{#1}}

\def\1{\bm{1}}

\DeclareMathAlphabet{\mathsfit}{\encodingdefault}{\sfdefault}{m}{sl}
\SetMathAlphabet{\mathsfit}{bold}{\encodingdefault}{\sfdefault}{bx}{n}

\DeclareMathOperator*{\argmax}{arg\,max}
\DeclareMathOperator*{\argmin}{arg\,min}

\definecolor{cvprblue}{rgb}{0.21,0.49,0.74}
\usepackage[pagebackref,breaklinks,colorlinks,citecolor=cvprblue]{hyperref}
\usepackage[capitalize]{cleveref}
\usepackage{multirow}
\crefname{section}{Sec.}{Secs.}
\Crefname{section}{Section}{Sections}
\Crefname{table}{Table}{Tables}
\crefname{table}{Tab.}{Tabs.}
\def\paperID{9054} 
\def\confName{CVPR}
\def\confYear{2024}
\usepackage{xspace}
\makeatletter\renewcommand\paragraph{\@startsection{paragraph}{4}{\z@}
	{.25em \@plus1ex \@minus.2ex}{-.5em}{\normalfont\normalsize\bfseries}}\makeatother
\title{L2B: Learning to Bootstrap Robust Models for Combating Label Noise}

\author{Yuyin Zhou$^{1}$\textsuperscript{*} \quad Xianhang Li$^{1}$\textsuperscript{*} \quad Fengze Liu$^{2}$ \quad Qingyue Wei$^{5}$ \\ \quad Xuxi Chen$^{3}$ \quad Lequan Yu$^{4}$ \quad Cihang Xie$^{1}$  \quad Matthew P. Lungren$^{5}$ \quad Lei Xing$^{5}$\vspace{0.3em} \\
\small $^{*}$equal contribution \vspace{.5em} \\
{\normalsize $^1$University of California, Santa Cruz} \quad
{\normalsize $^2$Johns Hopkins University} \quad 
{\normalsize $^3$UT Austin}\quad \\
{\normalsize $^4$The University of Hong Kong}\quad 
{\normalsize $^5$Stanford University}
}

\begin{document}
\maketitle
\begin{abstract}
Deep neural networks have shown great success in representation learning. However, when learning with noisy labels (LNL), they can easily overfit and fail to generalize to new data. This paper introduces a simple and effective method, named Learning to Bootstrap (L2B), which enables models to bootstrap themselves using their own predictions without being adversely affected by erroneous pseudo-labels.  It achieves this by dynamically adjusting the importance weight between real observed and generated labels, as well as between different samples through meta-learning. Unlike existing instance reweighting methods, the key to our method lies in a new, versatile objective that enables implicit relabeling concurrently, leading to significant improvements without incurring additional costs. 

L2B offers several benefits over the baseline methods.  It yields more robust models that are less susceptible to the impact of noisy labels by guiding the bootstrapping procedure more effectively. It better exploits the valuable information contained in corrupted instances by adapting the weights of both instances and labels. Furthermore, L2B is compatible with existing LNL methods and delivers competitive results spanning natural and medical imaging tasks including classification and segmentation under both synthetic and real-world noise. Extensive experiments demonstrate that our method effectively mitigates the challenges of noisy labels, often necessitating few to no validation samples, and is well generalized to other tasks such as image segmentation. This not only positions it as a robust complement to existing LNL techniques but also underscores its practical applicability. The code and models are available at~\url{https://github.com/yuyinzhou/l2b}.
\end{abstract}    
\section{Introduction}
\label{sec:intro}

In computer vision, deep learning has made significant strides, especially when provided with extensive, high-quality datasets. However, the persistent issue of label noise in real-world datasets, which stems from factors such as inter-observer variability, human annotation errors, and adversarial rival, can significantly undermine performance~\cite{nettleton2010study}. 
As the size of datasets for deep learning continues to grow, the impact of label noise may become more significant. Understanding and addressing label noise is crucial for improving the accuracy and reliability of deep learning models~\cite{liu2020early,wang2020learning,zheng2021meta,Yao_2021_CVPR,Zhu_2021_CVPR,wu2021ngc,zhou2021learning}.

Existing learning with noisy labels (LNL) methods, such as~\cite{patrini2017making,goldberger2017training}, focus on loss correction by estimating a noise corruption matrix, which is often challenging and involves assumptions~\cite{xia2019anchor,liu2015classification,hendrycks2018using}. Recent research like~\cite{jiang2018mentornet,han2018co,yu2019does} primarily targets identifying and utilizing clean samples within noisy datasets, frequently treating low-loss samples as clean~\cite{arpit2017closer}. Unlike approaches that discard noisy examples, meta-learning methods~\cite{ren2018learning,shu2019meta} assign adaptive weights to each sample, with noisier ones receiving lower weights. However, this may compromise performance in high-noise scenarios by neglecting or underweighting portions of the training data.
To better utilize corrupted samples, several studies have focused on using network predictions, or pseudo-labels~\cite{lee2013pseudo}, to recalibrate labels~\cite{reed2014training,tanaka2018joint,song2019selfie,yi2019probabilistic,arazo2019unsupervised}. The bootstrapping loss method~\cite{reed2014training} is notable for using pseudo-labels in training targets, countering noisy sample effects. However, the static weight of pseudo-labels can lead to overfitting and inadequate label correction~\cite{arazo2019unsupervised}. Addressing this, Arazo et al.~\cite{arazo2019unsupervised} developed a dynamic bootstrapping method that adjusts the balance between actual and pseudo-labels using a mixture model.

In contrast to prior works that individually reweight labels or instances, our paper introduces a simple and effective approach to concurrently adjust both, elegantly unified under a meta-learning framework. We term our method as \textbf{L}earning to \textbf{B}ootstrap (\textbf{L2B}), as our goal is to enable the network to self-boost its capabilities by harnessing its own predictions in combating label noise.
Specifically, L2B introduces a new, versatile loss that allows dynamically adjusting the balance between true and pseudo labels, as well as the weights of individual samples. This adjustment is based on performance metrics from a separate, clean validation set within a meta-network framework. Unlike previous bootstrapping loss methods~\cite{reed2014training,arazo2019unsupervised,zhang2020distilling}, which reallocate labels through a fixed weighted combination of pseudo and true labels, L2B offers greater flexibility. It uniquely does not limit the weights to sum to one, allowing for more nuanced reweighting across all instances and labels. Furthermore, we empirically show that meta-learning algorithms' need for a clean validation set can be removed by dynamically creating an online meta set from the training data using a Gaussian mixture model~\cite{permuter2006study}. This not only enhances our method's practicality but also facilitates its integration with current LNL techniques like DivideMix~\cite{li2020dividemix}, UniCon~\cite{karim2022unicon}, and C2D~\cite{zheltonozhskii2022contrast}. Consequently, L2B attains superior results without relying on a validation set.

In addition, we theoretically prove that our formulation, which reweights different loss terms, can be reduced to the original bootstrapping loss and therefore conducts an implicit relabeling instead. Through a meta-learning process, L2B achieves significant improvements (e.g., \textbf{+8.9\%} improvement on CIFAR-100 with 50\% noise) compared with the instance reweighting baseline with no extra cost.
This versatile bootstrapping procedure of L2B presents a simple and effective plug-in compatible with existing LNL methods.
Our comprehensive tests across both natural and medical image datasets such as CIFAR-10, CIFAR-100, Clothing 1M, and ISIC2019, covering various types of label noise and recognition tasks, highlight L2B's superiority over contemporary label correction and meta-learning techniques.

\section{Related Works}
\paragraph{Learning with noisy labels.} Various approaches have been proposed to tackle the challenge of training models with noisy labeled data.  Noisy detection approaches~\cite{hendrycks2018using,han2018co,zhang2022learning,yang2021estimating} focus on identifying and reducing the influence of noisy samples to mitigate label inaccuracies. Label correction strategies~\cite{reed2014training,li2020dividemix,zhang2020distilling} aim to refine pseudo labels to better match true labels. Within this domain, one group emphasizes robust representation learning through unsupervised contrastive learning~\cite{li2021learning,ghosh2021contrastive,zheltonozhskii2022contrast,karim2022unicon}, while another significant group employs meta-learning, using a subset of clean data for optimization guidance~\cite{ren2018learning,shu2019meta,xu2021faster,li2019learning,wu2021learning,zheng2021meta,zhang2020distilling}.
 
\paragraph{Explicit relabeling.}
Existing works propose to directly identify noisy samples and relabel them through estimating the noise transition matrix ~\cite{xia2019anchor,yao2020dual,goldberger2017training,patrini2017making} or modeling noise by graph models or neural networks~\cite{xiao2015learning,vahdat2017toward,veit2017learning,lee2018cleannet}. Patrini \emph{et al.}~\cite{patrini2017making} and  Hendrycks \emph{et al.}~\cite{hendrycks2018using} estimate the label corruption matrix to directly correct the loss function. 
However, these methods usually require assumptions about noise modeling.
For instance, Hendrycks \emph{et al.}~\cite{hendrycks2018using} assume that the noisy label is only dependent on the true label and
independent of the data.
Another line of approaches proposes to leverage the network prediction (pseudo-labels) for explicit relabeling~\cite{tanaka2018joint,yi2019probabilistic,han2019deep,reed2014training,Ortego_2021_CVPR}. 
However, using a uniform weight for all samples, as in~\cite{reed2014training}, can exacerbate the influence of noisy data, impeding effective label correction. Semi-supervised LNL techniques~\cite{li2020dividemix,zhang2020distilling} segment training data into labeled ``clean samples'' and unlabeled noisy sets, subsequently relabeled using pseudo-labels. To bolster the reliability of these pseudo-labels, unsupervised contrastive learning approaches are employed ~\cite{li2021learning,ghosh2021contrastive,zheltonozhskii2022contrast,karim2022unicon}.

\paragraph{Instance reweighting.}
To counteract the adverse effects of corrupted examples, various strategies focus on reweighting or selecting training instances to minimize the influence of noisy samples~\cite{jiang2018mentornet,ren2018learning,fang2020rethinking}. 
Based on the observation that deep neural networks tend to learn simple patterns first before fitting label noise~\cite{arpit2017closer},
many methods treat samples with small loss as clean ones~\cite{jiang2018mentornet,shen2019learning,han2018co,yu2019does,Wei_2020_CVPR}.
Rather than directly selecting clean examples for training, meta-learning techniques~\cite{ren2018learning,shu2019meta,xu2021faster} adjust instance weights, and curriculum learning~\cite{jiang2018mentornet} sequences them by noise levels. Such strategies enhance robustness in medical imaging~\cite{xue2019robust,mirikharaji2019learning}, but overlooking training subsets can affect performance in high-noise scenarios.

\paragraph{Meta-learning.} 
Meta-learning methods~\cite{ren2018learning,shu2019meta,xu2021faster,li2019learning,wu2021learning,zheng2021meta,zhang2020distilling} use a small clean validation set to optimize model weights and hyper-parameters. Techniques include instance reweighting~\cite{ren2018learning,shu2019meta,xu2021faster}, which involves bi-level optimization for determining training sample contributions. Another line of works view label correction as a separate meta-process~\cite{wu2021learning,zheng2021meta,zhang2020distilling}. Meta-learning has also been utilized to prevent overfitting to noisy labels~\cite{li2019learning}. 
Recently, meta-learning approaches has also been modified for other purposes. For instance, CMW-Net~\cite{shu2023cmw} adaptively generates sample weight based on the intrinsic bias characteristics of different sample classes.
DMLP~\cite{tu2023learning} combines self-supervised representation learning and a linear meta-learner for label correction.

Different from the aforementioned approaches which separately handle instance reweighting and label reweighting, we introduce a generic learning objective that concurrently meta-learns per-sample loss weights while implicitly relabeling the training data.

\section{Methodology}
\label{sec:method}
\subsection{Preliminary}
Given a set of $N$ training samples, i.e., $\mathcal{D}_{tra} = \{(x_i, y_i)| i = 1,...,N\}$, where $x_i \in \mathbb{R}^{W\times H}$ denotes the $i$-th image and $y_i$ is the observed noisy label. 
In this work, we also assume that there is a small unbiased and clean validation set $\mathcal{D}_{val} = \{(x^v_i, y^v_i)|i= 1,...,M\}$ and $M \ll N$, where the superscript $v$ denotes the validation set. 
Let $\mathcal{F} (:, \theta)$ denote the neural network model parameterized by $\theta$. 
Given an input-target pair $(x, y)$, we consider the loss function of
$\mathcal{L}(\mathcal{F} (x,\theta), y)$ (e.g., cross-entropy loss) to minimize during the training process.
Our goal in this paper is to properly utilize the small validation set $\mathcal{D}_{val}$ to guide the model training on $\mathcal{D}_{tra}$, for reducing the negative effects brought by the noisy annotation.

\begin{figure*}[t]
\begin{center}
\includegraphics[width=\textwidth]{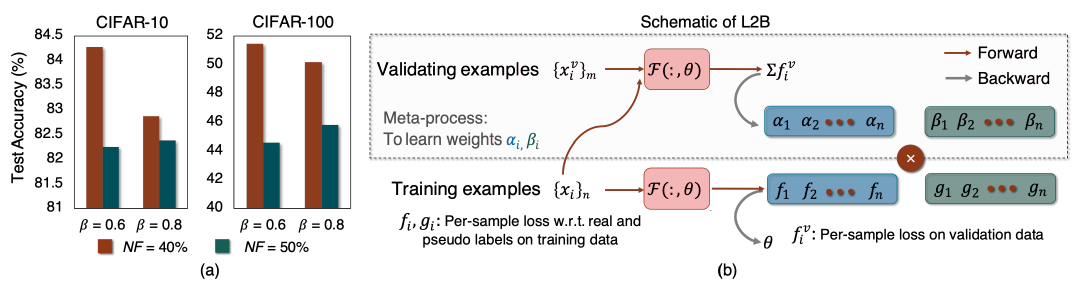}
\end{center}
\vspace{-1em}
  \caption{(a) The original bootstrapping loss~\cite{reed2014training} is sensitive to the reweighting hyper-parameter $\beta$. Under different noise levels, the optimal $\beta$ is different (\emph{NF} stands for noise fraction). (b) Schematic description of our Learning to Bootstrap (i.e., \textbf{L2B}) method. The reweighting hyper-parameters are learned in a meta-process.}
\vspace{-.5em}
\label{fig:overview}
\end{figure*}

To establish a more robust training procedure, ~\cite{reed2014training} proposed the bootstrapping loss to enable the learner to ``disagree'' with the original training label, and effectively re-label the data during the training. Specifically, the training targets will be generated using a convex combination of training labels and predictions of the current model (i.e., pseudo-labels~\cite{lee2013pseudo}), for purifying the training labels. %
Therefore, for a $L$-class classification problem, the loss function for optimizing $\theta$ can be derived as follows:
\begin{equation}
\label{Eqn:pseudo_gt}
y_i^\textup{pseudo} = \argmax_{l=1,..,L} \mathcal{P} (x_i,\theta),
\end{equation}
\begin{equation}
\label{Eqn:reed_hard_loss}
\theta^* = \argmin_\theta \sum_{i=1}^N \mathcal{L}(\mathcal{F} (x_i,\theta), \beta y_i^\textup{real} + (1 - \beta) y_i^\textup{pseudo}),
\end{equation}
where $\beta$ is used for balancing the weight between the real labels and the pseudo-labels. $\mathcal{P} (x_i,\theta)$ is the model output. $y^\textup{real}$ and $y^\textup{pseudo}$ denote the observed label and the pseudo-label respectively.
However, in this method, $\beta$ is manually selected and fixed for all training samples, which does not prevent fitting the noisy ones and can even lead to low-quality label correction~\cite{arazo2019unsupervised}.
Moreover, we observe that this method is quite sensitive to the selection of the hyper-parameter $\beta$.
For instance, as shown in Figure~\ref{fig:overview}(a), even a similar $\beta$ selection (i.e., $\beta=0.6$ vs. $\beta=0.8$) behaves differently under disparate noise levels, making the selection of $\beta$ even more intractable.
Another limitation lies in that ~\eqref{Eqn:reed_hard_loss} treats all examples as equally important during training, which could easily cause overfitting for biased training data.

\subsection{Learning to Bootstrap through Meta-Learning}
To address these above challenges, in this paper, we aim to learn to bootstrap the model by conducting a joint label reweighting and instance reweighting.
To achieve this, we propose to generate meta-learned weights for guiding our main learning objective: 
\begin{equation}
\label{eq:theta_star}
\begin{split}
\theta^*(\boldsymbol{\alpha}, \boldsymbol{\beta})  = \argmin_\theta &\sum_{i=1}^N \alpha_i \mathcal{L}(\mathcal{F} (x_i,\theta), y_i^\textup{real}) \\
&+ \beta_i \mathcal{L}(\mathcal{F} (x_i,\theta), y_i^\textup{pseudo}),
\end{split}
\end{equation}
with $\{\alpha_i, \beta_i\}_{i=1}^N$ being the balance weights. 
Here we note that this new learning objective can be regarded as a general form of the original bootstrapping loss, as ~\eqref{eq:theta_star} can be reduced to ~\eqref{Eqn:reed_hard_loss} when $\alpha_i + \beta_i = 1$ given that $\mathcal{L}(\cdot)$ is the cross-entropy loss (see details in Appendix~\ref{B.1}).
By relaxing this constraint such that $\boldsymbol{\alpha}, \boldsymbol{\beta} \ge \boldsymbol{0}$, we can see that the optimization of ~\eqref{eq:theta_star} not only allows the main learner to explore the optimal combination between the two loss terms but also concurrently adjust the contribution of different training samples.
In addition, compared with ~\eqref{Eqn:reed_hard_loss}, the optimization of ~\eqref{eq:theta_star} does not rely on explicitly generating new training targets (i.e., $\beta y_i^\textup{real} + (1 - \beta) y_i^\textup{pseudo}$), but rather conducts implicit relabeling during training by reweighting different loss terms.
We note that the key to L2B is that the sum of $\alpha_i$ and $\beta_i$ need not be 1, which results in \textbf{+8.9\%} improvement on CIFAR-100 with 50\% noise (Section~\ref{sec:performance}).

Note that this form is also similar to self-distillation in~\cite{li2017learning}. But different from~\cite{li2017learning} where the weights are determined by heuristics, our weights $\boldsymbol{\alpha}, \boldsymbol{\beta}$ are meta-learned based on its performance on the validation set $\mathcal{D}_{val}$, that is
\begin{equation}\label{eq:weights_star}
\boldsymbol{\alpha}^*, \boldsymbol{\beta}^* = \argmin_{\boldsymbol{\alpha}, \boldsymbol{\beta} \ge \boldsymbol{0}}\frac{1}{M} \sum_{i=1}^M\mathcal{L}(\mathcal{F} (x_i^v,\theta^*(\boldsymbol{\alpha}, \boldsymbol{\beta})), y_i^v).
\end{equation}
It is necessary to constrain $\alpha_i, \beta_i \ge 0$ for all $i$ to avoid potential unstable training~\cite{ren2018learning}.
Both the meta learner (i.e., ~\eqref{eq:weights_star}) and the main learner (i.e., ~\eqref{eq:theta_star}) are optimized concurrently, which allows the model to maximize the performance on the clean validation set $\mathcal{D}_{val}$ by adjusting the importance weights of the observed and the pseudo-labels in a differentiable manner.

\paragraph{Online Approximation.}
For each step $t$ at training, a mini-batch of training examples $\{(x_i,y_i),1\leq i\leq n\} $ with $n \ll N$ is sampled to estimate a temporary adjustment to the parameters based on the descent direction of the loss function. 
For simplicity, let $f_{i}(\theta)$ denote $\mathcal{L}(\mathcal{F} (x_i,\theta), y_i^\textup{real})$ and $g_{i}(\theta)$ denote $\mathcal{L}(\mathcal{F} (x_i,\theta), y_i^\textup{pseudo})$ in the following sections.
Given any $\boldsymbol{\alpha},\boldsymbol{\beta}$, we use 
\begin{equation}\label{eq:theta_hat}
\hat{\theta}_{t+1} = \theta_t - \lambda \nabla (\sum_{i=1}^n \alpha_i \ f_{i}(\theta)+\beta_i\ g_{i}(\theta))\Big |_{\theta=\theta_t}
\end{equation}
to approach the solution of ~\eqref{eq:theta_star}. Here $\lambda$ is the step size. We then estimate the corresponding optimal $\boldsymbol{\alpha},\boldsymbol{\beta}$ as
\begin{equation}\label{eq:alpha_t}
\boldsymbol{\alpha}_t^*, \boldsymbol{\beta}_t^* = \argmin_{\boldsymbol{\alpha}, \boldsymbol{\beta} \ge \boldsymbol{0}}\frac{1}{M} \sum_{i=1}^M f_i^v(\hat{\theta}_{t+1}).
\end{equation}

However, directly solving for ~\eqref{eq:alpha_t} at every training step requires too much computation cost. To reduce the computational complexity, we apply one step gradient descent of $\boldsymbol{\alpha}_t, \boldsymbol{\beta}_t$ on a mini-batch of validation set $\{(x_i^v,y_i^v),1\leq i \leq m\}$ with $m\leq M$ as an approximation. Specifically,
\begin{equation}\label{eq:alpha_beta}
(\alpha_{t,i},\beta_{t,i})= -\eta\nabla(\sum_{i=1}^m f_i^v(\hat{\theta}_{t+1}))\Big |_{\alpha_i=0,\beta_i=0},
\end{equation}
where $\eta$ is the step size for updating $\boldsymbol{\alpha},\boldsymbol{\beta}$. To ensure that the weights are non-negative, we apply the following rectified function:
\begin{equation}\label{eq:rectified_weights}
\Tilde{\alpha}_{t,i}=\text{max}(\alpha_{t,i},0),\ \Tilde{\beta}_{t,i}=\text{max}(\beta_{t,i},0).
\end{equation}
To stabilize the training process, we also normalize the weights in a single training batch so that they sum up to one: 
\begin{equation}\label{eq:alpha_norm}
\Tilde{\alpha}_{t,i}=\frac{\Tilde{\alpha}_{t,i}}{\sum_{i=1}^n \Tilde{\alpha}_{t,i}+\Tilde{\beta}_{t,i}},\ \Tilde{\beta}_{t,i}=\frac{\Tilde{\beta}_{t,i}}{\sum_{i=1}^n \Tilde{\alpha}_{t,i}+\Tilde{\beta}_{t,i}}.
\end{equation}
Finally, we estimate $\theta_{t+1}$ based on the updated $\boldsymbol{\alpha}_{t},\boldsymbol{\beta}_{t}$ so that $\theta_{t+1}$ can consider the meta information included in $\boldsymbol{\alpha}_{t},\boldsymbol{\beta}_{t}$:
\begin{equation}\label{eq:theta_t+1}
\theta_{t+1} = \theta_t - \lambda \nabla (\sum_{i=1}^n \Tilde{\alpha}_{t,i} \ f_{i}(\theta)+\Tilde{\beta}_{t,i}\ g_{i}(\theta))\Big |_{\theta=\theta_t}.
\end{equation}

See Appendix~\ref{B.2} for detailed calculation of the gradient in ~\eqref{eq:theta_t+1}. 
A schematic description of our Learning to Bootstrap algorithm is illustrated in Figure~\ref{fig:overview}(b) and the overall optimization procedure can be found in Algorithm~\ref{alg:L2B}.

\begin{algorithm}[t!]
\caption{Learning to Bootstrap}
\label{alg:L2B}
\begin{algorithmic}[1]
\REQUIRE $\theta_0$, $\mathcal{D}_{tra}$, $\mathcal{D}_{val}$, $n$, $m$, $L$
\ENSURE $\theta_T$
\FOR{$t=0$ ... $T-1$}
\STATE $\{x_i, y_i\} \gets$ \text{SampleMiniBatch}($\mathcal{D}_{tra}$, $n$)
\STATE $\{x_i^v, y_i^v\} \gets$ \text{SampleMiniBatch}($\mathcal{D}_{val}$, $m$)
\STATE For the $i$-th sample of $\mathcal{D}_{tra}$, compute $y^\textup{pseudo}_i = \argmax_{l=1,..,L} \mathcal{P} (x_i,\theta_t)$ 
\STATE Learnable weights $\boldsymbol{\alpha}$, $\boldsymbol{\beta}$
\STATE Compute training loss $l_f \gets \sum_{i=1}^n \alpha_i f_i(\theta_t) + \beta_i g_i(\theta_t)$
\STATE $\hat{\theta}_{t + 1} \gets \theta_t - \lambda \nabla l_f\Big |_{\theta=\theta_t}$
\STATE Compute validation loss $l_g \gets \frac{1}{m} \sum_{i=1}^m f^v_i(\hat{\theta}_{t + 1})$
\STATE $(\boldsymbol{\alpha}_{t},\boldsymbol{\beta}_{t})\gets-\eta\nabla l_g \Big |_{\boldsymbol{\alpha}=\boldsymbol{0},\boldsymbol{\beta}=\boldsymbol{0}}$
\STATE $\Tilde{\alpha}_{t,i}\gets\text{max}(\alpha_{t,i},0),\ \Tilde{\beta}_{t,i}\gets\text{max}(\beta_{t,i},0)$
\STATE $\Tilde{\alpha}_{t,i}\gets\frac{\Tilde{\alpha}_{t,i}}{\sum_{i=1}^n \Tilde{\alpha}_{t,i}+\Tilde{\beta}_{t,i}},\ \Tilde{\beta}_{t,i}\gets\frac{\Tilde{\beta}_{t,i}}{\sum_{i=1}^n \Tilde{\alpha}_{t,i}+\Tilde{\beta}_{t,i}}$
\STATE Apply learned weights $\boldsymbol{\alpha}, \boldsymbol{\beta}$ to reweight the training loss as $\hat{l}_f \gets \sum_{i=1}^n \Tilde{\alpha}_{t,i} f_i(\theta_t) + \Tilde{\beta}_{t,i} g_i(\theta_t)$
\STATE  $\theta_{t + 1} \gets \theta_t - \lambda \nabla \hat{l}_f\Big |_{\theta=\theta_t}$
\ENDFOR
\end{algorithmic}
\end{algorithm}

\begin{table*}[htb]
\footnotesize
    \centering
    \caption{
        Comparison in test accuracy (\%) with the baseline methods on CIFAR-10/100 datasets with symmetric noise.
        }\vspace{-0.5em}
\label{tab:baseline}
    \resizebox{0.85\textwidth}{!}{
    \begin{tabular}{l |c  c  c c| c c cc|c c c c}
        \toprule[0.1em]
        Dataset                                                                &      &                  \multicolumn{3}{c|}{CIFAR-10~~~~~~~~~~~~~~~~~~}                  &                    \multicolumn{4}{c|}{CIFAR-100}       &                    \multicolumn{4}{c}{ISIC}               \\ 
        \midrule[0.08em]
        Method/Noise                                                        ratio      & 20\%             &     30\%      &     40\%      &     50\%      &       20\%       &       30\%       &     40\%      &     50\%    &       20\%       &       30\%       &     40\%      &     50\%   \\ \hline
       {Cross-Entropy (CE)}                                          & 86.9             &     84.9      &     83.3      &     81.3      &       59.6       &       52.2       &     49.2      &     44.4     & 79.4&77.5 &75.3 &73.7  \\\hline
       {Bootstrap~\cite{reed2014training}}           & 85.2             &     84.8      &     82.9      &  79.2      &      61.8       &       54.2       &    50.2      &     45.8   &80.8 &77.7 &75.7 &74.8   \\\hline
      {L2RW~\cite{ren2018learning}}           & 90.6             &     89.0      &     86.6     &     85.3      &       67.8       &       63.8       &     59.7      &    55.6     &80.1 &77.7 &76.3 &74.1 \\\hline
        \rowcolor{mygray}{L2B (Ours)}           & \textbf{92.2}             &     \textbf{90.7}      &   \textbf{89.9}      &     \textbf{88.5}      &       \textbf{71.8}       & \textbf{69.5}       &  \textbf{67.3}     &    \textbf{64.5}  &\textbf{81.1} &\textbf{80.2} &\textbf{78.6} &\textbf{76.8}   \\
        \bottomrule[0.1em]
     \end{tabular}
     }
\end{table*}

\subsection{Convergence Analysis} 
\label{sec:convergence}
In proposing ~\eqref{eq:theta_star}, we show that with the first-order approximation of $\boldsymbol{\alpha}, \boldsymbol{\beta}$ in ~\eqref{eq:alpha_beta} and some mild assumptions, our method guarantees to convergence to a local minimum point of the validation loss, which yields the best combination of $\boldsymbol{\alpha}, \boldsymbol{\beta}$. 
Details of the proof are provided in Appendix~\ref{B.3}.

\section{Experiments}
\subsection{Datasets}

\paragraph{CIFAR-10 \& CIFAR-100.}~\label{sec:data} Both CIFAR-10 and CIFAR-100 contain 50K training images and 10K test images of size 32 × 32.
Following previous works~\cite{tanaka2018joint,kim2019nlnl,li2020dividemix}, we experimented with both \emph{symmetric} and \emph{asymmetric} label noise.
In our method, we used 1,000 clean images in the validation set $\mathcal{D}_{val}$ following~\cite{jiang2018mentornet,ren2018learning,shu2019meta,hendrycks2018using,zheng2021meta}. 

\paragraph{ISIC2019.} Following~\cite{xue2019robust}, we also evaluated our algorithm on a medical image dataset, i.e., skin lesion classification data, under different symmetric noise levels. 
Our experiments were conducted on the 25,331 dermoscopic images of the 2019 ISIC Challenge\footnote{\url{https://challenge2019.isic-archive.com/data.html}}, where we used 20400 images as the training set $\mathcal{D}_{tra}$, 640 images as the validation set $\mathcal{D}_{val}$, and tested on 4291 images.

\paragraph{Clothing 1M.} We evaluate on real-world noisy dataset, Clothing 1M~\cite{xiao2015learning}, which has 1 million training images collected from online shopping websites with labels generated
from surrounding texts.
In addition, the Clothing 1M also provides an official validation set of 14,313  images and a test set of
10,526 images.

\subsection{Implementation Details}
For all CIFAR-10 and CIFAR-100 comparison experiments, we used an 18-layer PreActResNet~\cite{he2016identity} as the baseline network following the setups in~\cite{li2020dividemix}, unless otherwise specified.
The model was trained using SGD with a momentum of
0.9, a weight decay of 0.0005, and a batch size of 256 for CIFAR-100 and 512 for CIFAR-10. 
The network was trained from scratch for 300 epochs. We
set the learning rate as 0.15 initially with a cosine annealing decay. Following~\cite{li2020dividemix}, we set the warm up
period as 10 epochs for both CIFAR-10 \& CIFAR-100. 
The optimizer and the learning rate schedule remained the same for both the main and the meta model. 
Gradient clipping is applied to stabilize training.
All experiments were conducted with one V100 GPU, except for the experiments on Clothing 1M which were conducted with one RTX A6000 GPU.

For ISIC2019 experiments, we used ResNet-50 with ImageNet pretrained weights. A batch size of 64 was used for training with an initial learning rate of 0.01. The network was trained for 30 epochs in total with the warmup period as 1 epoch.
All other implementation details remained the same as above.
For Clothing 1M experiments, we used an ImageNet pre-trained 18-layer ResNet~\cite{he2016identity} as our baseline. We finetuned the network with a learning rate of 0.005 for 300 epochs. The model was trained using SGD with a momentum of 0.9, a weight decay of 0.0005, and a batch size of 256. Following~\cite{li2020dividemix}, to ensure the labels (noisy) were balanced, for each epoch, we sampled 250 mini-batches from the training data.

\begin{table*}[t]
\footnotesize
    \centering
    \caption{
        Comparison in test accuracy (\%) with state-of-the-art methods on CIFAR-10/100 datasets with symmetric noise.
        }\vspace{-0.5em}
    \setlength\tabcolsep{10pt} \renewcommand{\arraystretch}{1.3}
    \resizebox{0.73\textwidth}{!}{
    \begin{tabular}{l |c  c  c c| c c cc}
        \toprule[0.1em]
        Dataset                                                                &      &                  \multicolumn{3}{c|}{CIFAR-10~~~~~~~~~~~~~~~~~~}                  &                    \multicolumn{4}{c}{CIFAR-100}                     \\ 
        \midrule[0.08em]
        Method/Noise                                                        ratio      & 20\%             &     50\%      &     80\%      &     90\%      &       20\%       &       50\%       &     80\%      &     90\%      \\ \hline
                                                                        
        {Co-teaching$+$~\cite{yu2019does}}                        & 89.5             &     85.7      &     67.4      &     47.9      &       65.6       &       51.8       &     27.9      &     13.7      \\
                                                                                \hline
        {Mixup~\cite{zhang2018mixup}}                            & 95.6             &     87.1      &     71.6      &     52.2      &       67.8       &       57.3       &     30.8      &     14.6      \\
                                                                                \hline
        {PENCIL~\cite{yi2019probabilistic}}                      & 92.4             &     89.1      &     77.5      &     58.9      &       69.4       &       57.5       &     31.1      &     15.3      \\
                                                                                \hline
        {Meta-Learning~\cite{li2019learning}}                              & 92.9             &     89.3      &     77.4      &     58.7      &       68.5       &       59.2       &     42.4      &     19.5      \\
                                                                               \hline
        {M-correction~\cite{arazo2019unsupervised}}              & 94.0             &     92.0      &     86.8      &     69.1      &       73.9       &       66.1       &     48.2      &     24.3      \\
                                                                                \hline                                                                          
        {AugDesc~\cite{nishi2021augmentation}}                         & 96.3 & 95.4 & 93.8 & 91.9 & 79.5 & 77.2 & 66.4 & 41.2 \\
                                                                               \hline                                                                       
        {GCE~\cite{ghosh2021contrastive}}                         & 90.0 & 89.3 & 73.9 & 36.5 & 68.1 & 53.3& 22.1 & {8.9} \\
                                                                              \hline
                                                                               
        {Sel-CL+~\cite{li2022selective}}                         & 95.5 & 93.9 & 89.2& 81.9 & 76.5 & 72.4 & 59.6 & {48.8} \\
                                                                               \hline
        {MLC~\cite{zheng2021meta}} &   92.6        &   88.1        &  77.4         & 67.9        &   66.8           &   52.7           &  21.8        &  15.0         \\
        \hline
       {MSLC~\cite{wu2021learning}} & 93.4     & 89.9         & 69.8       &  56.1        &   72.5           &  65.4           &   24.3        & 16.7         \\
       \hline
                                                                               
       {MOIT+~\cite{ortego2021multi}}                         & 94.1 & 91.8 & 81.1 & 74.7 & 75.9 & 70.6 & 47.6 & {41.8} \\
                                                                                 \hline
     {TCL}~\cite{huang2023twin}                         &95.0 & 93.9 & 92.5 & 89.4 & 78.0 & 73.3 & 65.0 & 54.5\\
                                                                                 \hline 
     {DivideMix~\cite{li2020dividemix}}             & {96.1} &     {94.6}      &     {93.2}      &     76.0     &       {77.3}       &       {74.6}       &     60.2      &     31.5      \\
         \rowcolor{mygray} {L2B-DivideMix}       
          &96.1 &\textbf{95.4}  &\textbf{94.0} &\textbf{91.3 }
          &\textbf{77.9}  &\textbf{75.9}   &\textbf{62.2} &\textbf{35.8}

              \\
\hline
    {UniCon~\cite{karim2022unicon}}  &  96.0  &  {95.6}  &  {93.9}  &  {90.8}   &   {78.9}  &  {77.6}  &  {63.9}  &  {44.8}  \\ 
   \rowcolor{mygray} {L2B-UniCon}  
  &\textbf{96.5}&\textbf{95.8}&\textbf{94.7}&\textbf{92.8}
  &78.8&77.3&\textbf{67.6}&\textbf{49.6}    \\ 
    \hline
    {C2D~\cite{zheltonozhskii2022contrast}}            
    & {96.3} & 95.2 & 94.4 & 93.5 & 78.7 & 76.4 & 67.8 & {58.7} \\ 
   \rowcolor{mygray} {L2B-C2D}   
    &\textbf{96.7}&\textbf{95.6}&\textbf{94.8}&\textbf{94.4}
    &\textbf{80.1}&\textbf{78.1}&\textbf{69.6}&\textbf{60.7}
    \\
                        \bottomrule[0.1em]
    \end{tabular}}
    \label{tab:cifar_sym}
\end{table*}
\subsection{Performance Comparisons}
\label{sec:performance}
\begin{figure}
\begin{center}
\includegraphics[width=0.49\textwidth]{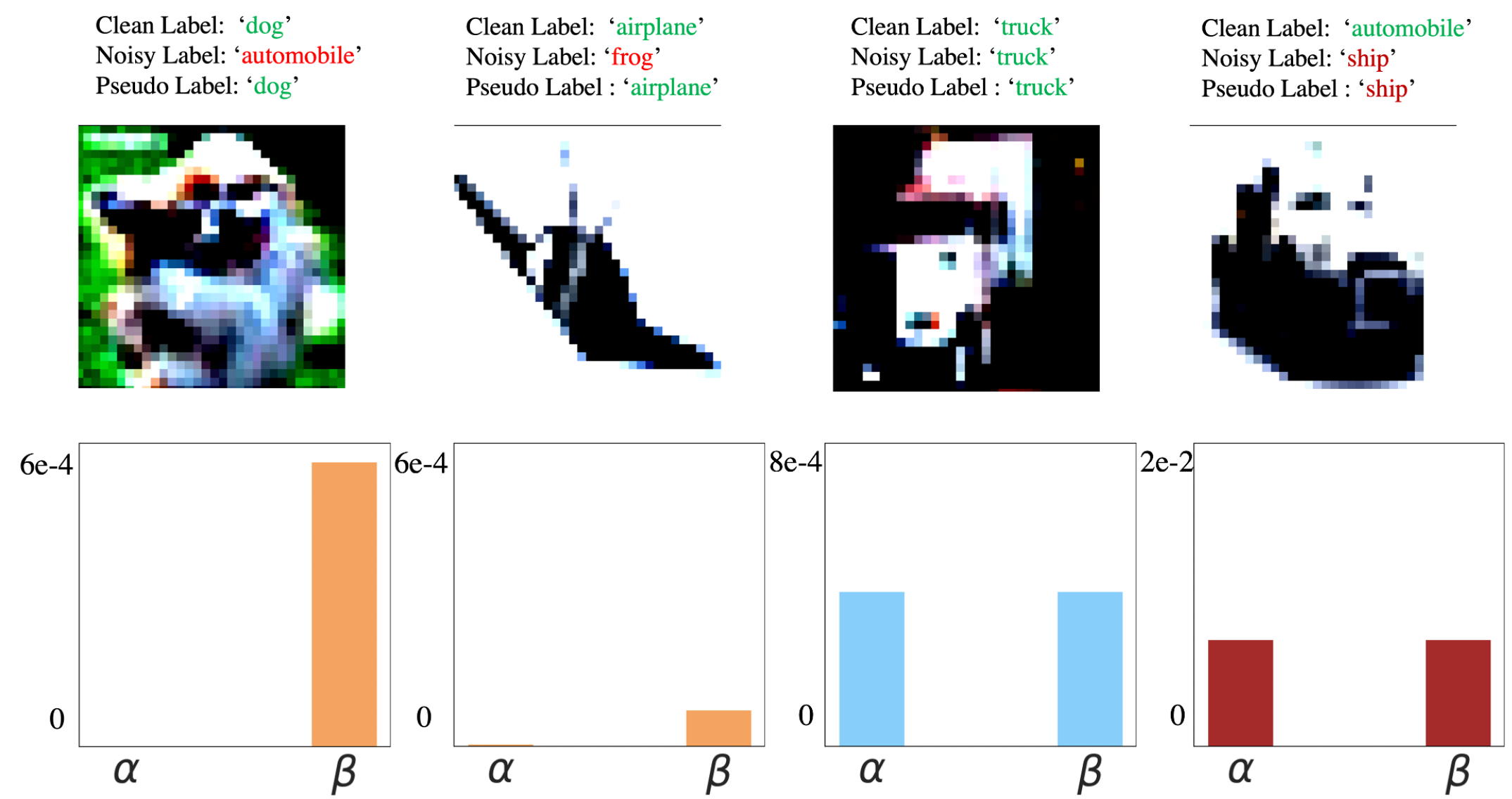}
\end{center}
\vspace{-1.em}
  \caption{Examples of $\alpha$ and $\beta$ on CIFAR-10 with asymmetric noise fraction of 20\%. When the estimated  pseudo label is of high-quality, i.e., the pseudo label is different from the noisy label but equal to the clean label, our model will automatically assign a much higher weight to $\beta$ than to $\alpha$ for corrupted training samples. When the pseudo label is equal to the noisy label (i.e., the two loss terms are equal to each other), $\alpha$ and $\beta$ are almost identical. 
  }
\label{fig:example_weight}
\vspace{-.5em}
\end{figure}

\paragraph{Efficacy of L2B.} We compare our method with different baselines: 1) Cross-Entropy (the standard training), 2) Bootstrap~\cite{reed2014training}, which modifies the training loss by generating new training targets
,  and 3) L2RW~\cite{ren2018learning}, which reweights different instances through meta-learning under different levels of symmetric labels noise ranging from $20\%\sim 50\%$.
To ensure a fair comparison, we report the best epoch for all comparison approaches.
All results are summarized in Table~\ref{tab:baseline}. 
Compared with the naive bootstrap method and the baseline meta-learning-based instance reweighting method L2RW, the performance improvement is substantial, especially under larger noise fraction, which suggests that using meta-learning to automatically bootstrap the model is more beneficial for LNL.
For example, on CIFAR-100, the accuracy improvement of our proposed L2B reaches $7.6\%$ and $8.9\%$ under 40\% and 50\% noise fraction, respectively. We also demonstrate a set of qualitative examples to illustrate how our proposed L2B benefits from the joint instance and label reweighting paradigm in Figure~\ref{fig:example_weight}. 
We can see that when the online estimated  pseudo label is of high-quality, i.e., the pseudo label is different from the noisy label but equal to the clean label, our model will automatically assign a much higher weight to $\beta$ for corrupted training samples. On the contrary, $\alpha$ can be near zero in this case. This indicates that our L2B algorithm will pay more attention to the correct pseudo label than the real noisy label when computing the losses.
In addition, we also show several cases where the online pseudo label has not yet been corrected and therefore is equal to the noisy label during the training process, where we can see that $\alpha$ and $\beta$ are almost identical under this circumstance since there will be no need to correct it. 
We note that by the end of the training, most noisy examples will be successfully corrected, leading to significantly different weighting of pesudo and noisy labels that will help rectify the training.

The relatively small values of $\alpha$ and $\beta$ are due to that we use a large batch size (i.e., 512) for CIFAR-10 experiments. By normalizing the weights in each training batch (see~\eqref{eq:alpha_norm}), the value of $\alpha$ and $\beta$ can be on the scale of $10^{-4}$.
\begin{table}
\captionof{table}{Comparison with 40\% asymmetric noise in test accuracy on the CIFAR-10 dataset.}\vspace{-0.5em}
\label{tab:aymmetric_noise}
\scriptsize
\centering
\resizebox{0.78\linewidth}{!}{
\tabcolsep=0.65cm
    \begin{tabular}[b]{c|c}
                \toprule[0.1em]
                Method   &Acc   \\
                 \midrule[0.08em]
               Cross-Entropy &85.0 \\
                F-correction~\cite{patrini2017making}  &87.2 \\
                M-correction~\cite{arazo2019unsupervised} &87.4     \\
                Chen~et al.~\cite{chen2019understanding} &88.6    \\
                P-correction~\cite{yi2019probabilistic} &88.5    \\
                REED~\cite{zhang2020decoupling} &92.3\\

                Tanaka~et al.~\cite{tanaka2018joint} &88.9    \\
                NLNL~\cite{kim2019nlnl} &89.9 \\
                JNPL~\cite{Kim_2021_CVPR} &90.7 \\
                DivideMix~\cite{li2020dividemix}&93.4\\

                \hline
                MLNT~\cite{li2019learning} &89.2    \\
                L2RW~\cite{ren2018learning} &89.2    \\
                MW-Net~\cite{shu2019meta} &89.7 \\
                MSLC~\cite{wu2021learning}   &91.6\\

                Meta-Learning~\cite{li2019learning} &88.6 \\
                Distilling~\cite{zhang2020distilling}&90.2\\

                \hline

                 \rowcolor{mygray}{L2B-Naive (Ours)} & 91.8    \\
                 
                 \rowcolor{mygray}{L2B-C2D (Ours)} & \textbf{94.0}    \\
                \bottomrule[0.1em]
    \end{tabular}
    }
\end{table}

\begin{table}[h!]
    \captionof{table}{Comparison with state-of-the-art methods in test accuracy (\%) on Clothing 1M. }\vspace{-0.5em}
\label{tab:clothing1M_comparison}
\centering
\scriptsize
\resizebox{0.89\linewidth}{!}{
\tabcolsep=0.6cm
\begin{tabular}{c|c}
    \toprule[0.1em]       
        Method                                       &  Acc (\%)   \\ \midrule[0.08em]
            Cross-Entropy                                 &  69.2                 \\ 
        M-correction~\cite{arazo2019unsupervised}    &  71.0                 \\ 
        PENCIL~\cite{yi2019probabilistic}            &  73.5                 \\ 
        DivideMix~\cite{li2020dividemix}             &  74.8                  \\ 
        Nested~\cite{chen2021boosting}   & 74.9  \\
        AugDesc~\cite{nishi2021augmentation} & 75.1 \\
        RRL~\cite{li2021learning}  &  74.9            \\ 
        GCE~\cite{ghosh2021contrastive}  &  73.3            \\ 
        C2D~\cite{zheltonozhskii2022contrast}  &  74.3             \\ \hline
        MLNT~\cite{li2019learning}               &  73.5                                 \\
        MLC ~\cite{zheng2021meta} &75.8                  \\ 
        MSLC ~\cite{wu2021learning} & 74.0                  \\ 
        Meta-Cleaner~\cite{zhang2019metacleaner}   &  72.5                 \\ 
        Meta-Weight~\cite{shu2019meta}  & 73.7 \\
        FaMUS~\cite{xu2021faster}  & 74.4 \\
        MSLG~\cite{algan2021meta}  &76.0 \\
        DISC~\cite{li2023disc} & 73.7 \\
        InstanceGM~\cite{garg2023instance} &74.4\\
        DivideMix+SNSCL~\cite{wei2023fine} & 75.3\\
        \midrule
        \rowcolor{mygray}\textbf{L2B-Naive (Ours)}                                     &  $\textbf{77.5}\pm0.2$   \\    
        \bottomrule[0.1em]
    \end{tabular}}
    \vspace{-1.2em}
\end{table}

\paragraph{Comparison with the state-of-the-arts.} We compare our method with SOTA methods on CIFAR 10 and CIFAR 100 in Table~\ref{tab:cifar_sym}. We demonstrate our L2B is compatible with existing LNL methods. When integrated with existing LNL methods like DivideMix~\cite{li2020dividemix}, UniCon~\cite{karim2022unicon}, C2D~\cite{zheltonozhskii2022contrast}, L2B consistently enhances performance across varying noise ratios on both datasets. Notably, L2B-C2D surpasses all competing methods in various settings, achieving 94.4\% and 60.7\% accuracy under the noise ratio of 90\% for CIFAR-10 and CIFAR-100. We also test our model with 40\% asymmetric noise and summarize the testing accuracy in Table~\ref{tab:aymmetric_noise}. Among all compared methods, we re-implement L2RW under the same setting and report the performance of all other competitors from previous papers including~\cite{kim2019nlnl,Kim_2021_CVPR,li2020dividemix}.Compared with previous meta-learning-based methods (\emph{e.g.,}~\cite{chen2019understanding}, ~\cite{zhang2020decoupling}), and other methods (\emph{e.g.,}~\cite{ren2018learning},~\cite{wu2021learning},~\cite{shu2019meta}), our L2B achieves superior results.

\begin{table}
    \captionof{table}{Segmentation performance comparison under noisy-supervision on PROMISE12. }\vspace{-0.5em}
\label{compare_noisy}
\centering
\resizebox{\linewidth}{!}{
\begin{tabular}{c|c|c|c}
\toprule[0.1em]
Method & Dice (\%)$\uparrow$ $\uparrow$ & HD (voxel)$\downarrow$& ASD (voxel)$\downarrow$  \\ \midrule[0.08em]
UNet++~\cite{zhou2018unet++} &73.74 	&11.63	&3.70	 \\ 
UNet++ meta &73.04		&17.06	&5.50\\ 
NL reweighting ~\cite{mirikharaji2019learning}  &76.64	 &8.33	&2.75	\\ 
Mix-up~\cite{zhang2017mixup}  &69.18  &13.25 &4.56	\\ 
\rowcolor{mygray}{L2B (Ours)} 
& $\bm{80.83}$  &$\bm{6.68}$ &$\bm{2.10}$\\ 
\bottomrule[0.1em]
\end{tabular}
}
\end{table}

\begin{table}[thb]
    \centering   
    \caption{ L2B for segmentation under different noise levels.}
    \resizebox{0.7\linewidth}{!}{
    \tabcolsep=0.8cm
    \begin{tabular}{c|c}
    \toprule[0.1em]
Method & Dice (\%)$\uparrow$ \\ \hline
baseline - $\text{L}_1$ &59.77 \\
\textbf{L2B} - $\text{L}_1$ &77.70 \\ \hline
baseline - $\text{L}_2$ &73.74 \\ 
\textbf{L2B} - $\text{L}_2$& 80.83 \\ \hline
baseline - $\text{L}_3$ &80.03 \\
\textbf{L2B} - $\text{L}_3$ &82.01 \\ 
\bottomrule[0.1em]
\end{tabular}}
\label{noisy level ab_new}
\end{table}

\begin{table*}\vspace{-0.5em}
\footnotesize
\centering
\caption{Ablation on size of validation data on CIFAR-10 and CIFAR-100 datasets.}
\vspace{-.5em}
\resizebox{0.9\linewidth}{!}{
\tabcolsep=0.3cm
\begin{tabular}{ccc|cccc|cccc}
\toprule
\multicolumn{2}{c}{} &  & \multicolumn{4}{c|}{CIFAR-10}  &\multicolumn{4}{c}{CIFAR-100} \\
&&Validation Size &20\% &50\% &80\% &90\% &20\% &50\% &80\% &90\% \\
\midrule
\multirow{4}{*}{{L2B-DivideMix}}
&&baseline & 96.1 & 94.6 & 93.2 & 76.0 & 77.3 & 74.6 & 60.2 & 31.5 \\
&&0       &  96.3 & 95.3 & 93.5 & 82.6 & 77.6 & 75.3 & 60.8 & 31.0 \\
&&500      & 96.1 & 95.3 & 93.8 & 91.1 & 78.2 & 75.3 & 62.5 & 34.0 \\
&&1000     & 96.1 & 95.4 & 94.0 & 91.3 & 77.9 & 75.9 & 62.2 & 35.8 \\
\midrule
\multirow{4}{*}{L2B-UniCon} 
&&baseline& 96.0 & 95.6 & 93.9 & 90.8 & 78.9 & 77.6 & 63.9 & 44.8 \\
&&0 & 96.4 & 95.6 & 94.2 & 92.5 & 78.7 & 77.4 & 68.0 & 48.6 \\
&&500 & 96.3 & 95.6 & 94.5 & 92.7 & 78.5 & 77.5 & 67.8 & 51.1 \\
&&1000 & 96.5 & 95.8 & 94.7 & 92.8 & 78.8 & 77.3 & 67.6 & 49.6 \\
\midrule
\multirow{4}{*}{L2B-C2D} 
&&baseline &96.4 & 95.3 & 94.4 & 93.5 & 78.7 & 76.4 & 67.8 & 58.7 \\
&&0 &96.4 & 95.6 & 94.9 & 93.7 & 79.1 & 77.8 & 68.5 & 60.3 \\
&&500 &96.6 & 95.5 & 94.9 & 94.0 & 79.5 & 77.9 & 69.0 & 60.8 \\
&&1000 &96.7 & 95.6 & 94.8 & 94.4 & 80.1 & 78.1 & 69.6 & 60.7 \\
\bottomrule
\end{tabular}}
\label{tab:validation-size}
\end{table*}

\paragraph{Generalization to real-world noisy labels.} 
We test L2B on Clothing 1M~\cite{xiao2015learning}, a large-scale dataset with real-world noisy labels. The results of all competitors are reported from published papers. As shown in Table~\ref{tab:clothing1M_comparison}, our L2B-Naive attains an average performance of 77.5\% accuracy from 3 independent runs with different random seeds, outperforming all competing methods.

\paragraph{Generalization to image segmentation}
L2B can be easily generalized for segmentation tasks. Specifically, the learnable weights $\boldsymbol{\alpha}$ and $\boldsymbol{\beta}$ are replaced with pixel-wise weight maps corresponding to noisy labels and pseudo labels (model predictions). L2B dynamically assigns these weight maps, adjusting for both noisy and pseudo labels to optimize the bootstrapping process via a meta-process. To assess L2B's performance in segmentation, we employed the PROMISE12 dataset~\cite{litjens2014evaluation} which contains 50 3D transversal T2-weighted MR images. Specifically, 40/10 cases were used for training/evaluation. 3 out of the 40 training cases are chosen randomly as the meta set. Following~\cite{soerensen2021deep,wang2021tripled}, we utilized 2D slices in the axial view for both training and testing. All images are resized to $144 \times 144$ and splits are randomized. Noisy labels used in Table~\ref{compare_noisy} were synthesized using random rotation, erosion, or dilation, achieving approximately a 60\% corruption ratio and an average Dice coefficient of $0.6206$. And visualizations of the corrupted noisy labels (shown in yellow) as well as the ground-truth (shown in red) are illustrated in Figure~\ref{fig:noisy_vis} in Appendix. As presented in Table~\ref{compare_noisy}, we compare our method with 1) UNet++~\cite{zhou2018unet++}, 2) UNet++ meta, which trains exclusively on the meta data, 3) NL reweighting~\cite{mirikharaji2019learning}, which only reweights the noisy labels, 4) Mix-up~\cite{zhang2017mixup}, a regularization based method. L2B outperforms others in all evaluation metrics of Dice, Hausdorff Distance (HD) and Average Surface Distance (ASD). Furthermore, we also investigate the robustness of our method by varying the noise level of the corrupted training set from $\{\text{L}_1, \text{L}_2, \text{L}_3\}$, where the average Dice coefficients are $\text{Dice}_{\text{L}_1}$ = $0.4148$, $\text{Dice}_{\text{L}_2}$ = $0.6206$, and $\text{Dice}_{\text{L}_3}$ = $0.8031$ (\emph{i.e.}, the corrupted ratios are around 60\% ($\text{L}_1$), 40\% ($\text{L}_2$), and 20\% ($\text{L}_3$)). At each noise level, we compare the baseline UNet++ which is directly trained on the noisy training data with our generalized L2B. As shown in Table~\ref{noisy level ab_new}, we report the averaged dice coefficient over 5 repetitions for each series of experiments. The standard deviation for all experiments is within 0.5\%. We could notice that while the noise level increases, performances of baseline drop from 80.03\% to 59.77\%, but performances of L2B only drop from 82.01\% to 77.70\% which indicates that our L2B is robust to different noisy levels and shows larger improvements under a much severer noisy situation.

\paragraph{Qualitative Results}
We also demonstrate a set of qualitative examples to illustrate how our proposed L2B benefits from the joint instance and label reweighting paradigm. 
In Figure~\ref{fig:example_weight}, we can see that when the estimated  pseudo label is of high-quality, i.e., the pseudo label is different from the noisy label but equal to the clean label, our model will automatically assign a much higher weight to $\beta$ for corrupted training samples. On the contrary, $\alpha$ can be near zero in this case. This indicates that our L2B algorithm will pay more attention to the pseudo label than the real noisy label when computing the losses.
In addition, we also show several cases where the pseudo label is equal to the noisy label, where we can see that $\alpha$ and $\beta$ are almost identical under this circumstance since the two losses are of the same value. 
Note that the relatively small values of $\alpha$ and $\beta$ are due to that we use a large batch size (i.e., 512) for CIFAR-10 experiments. By normalizing the weights in each training batch (see ~\eqref{eq:alpha_norm}), the value of $\alpha$ and $\beta$ can be on the scale of $10^{-4}$.

\subsection{Ablation Study}

\paragraph{On the importance of $\boldsymbol{\alpha}, \boldsymbol{\beta}$.} To understand why our proposed new learning objective can outperform previous meta-learning-based instance reweighting methods, we conduct the following analysis to understand the importance of hyper-parameter $\boldsymbol{\alpha}$ and $\boldsymbol{\beta}$ in our method. Specifically, we set $\boldsymbol{\alpha} = \boldsymbol{0}$ and $\boldsymbol{\beta} = \boldsymbol{0}$ respectively to investigate the importance of each loss term in ~\eqref{eq:theta_star}.
In addition, we also show how the restriction of $\alpha_i + \beta_i = 1$ (\eqref{Eqn:reed_hard_loss}) would deteriorate our model performance as follows. 

\begin{itemize}
   \item $\boldsymbol{\alpha} = \boldsymbol{0}$. As shown in Table~\ref{tab:ablation}, in this case, the performance even decreases compared with the baseline approach. 
   This is due to that when only pseudo-labels are included in the loss computation, the error which occurs in the initial pseudo-label will be reinforced by the network during the following iterations.
   \item $\boldsymbol{\beta} = \boldsymbol{0}$. From ~\eqref{eq:theta_star}, we can see that setting $\boldsymbol{\beta}$ as $\boldsymbol{0}$  is essentially equivalent to the baseline meta-learning-based instance reweighting method L2RW~\cite{ren2018learning}.
   In this case, the performance is largely improved compared to the baseline, but still inferior to our method, which jointly optimizes $\boldsymbol{\alpha}$ and $\boldsymbol{\beta}$.

   \item  $\boldsymbol{\alpha} + \boldsymbol{\beta} = \boldsymbol{1}$. We also investigate whether the restriction of $\boldsymbol{\alpha} + \boldsymbol{\beta} = \boldsymbol{1}$ is required for obtaining optimal weights during the meta-update, as in~\cite{zhang2020distilling}. 
   As shown in Table~\ref{tab:ablation}, L2B ($\boldsymbol{\alpha}, \boldsymbol{\beta} \ge \boldsymbol{0}$) consistently achieves superior results than L2B ($\boldsymbol{\alpha} + \boldsymbol{\beta} = \boldsymbol{1}$) under different noise levels on CIFAR-100. The reason may be the latter is only reweighting different loss terms, whereas the former not only explores the optimal combination between the two loss terms but also jointly adjusts the contribution of different training samples.
\end{itemize}

\begin{table}[]
\captionof{table}{Ablation of $\alpha,\beta$. L2B ($\alpha, \beta \ge 0$) consistently achieves superior results to L2B ($\alpha + \beta = 1$) under different noise levels on CIFAR-100.}\vspace{-0.5em}
\label{tab:ablation}
\centering
\resizebox{0.73\linewidth}{!}{
\tabcolsep=0.6cm
    \begin{tabular}{c|c|c}
                   \toprule[0.1em]
                    Method   &20\%   &40\%  \\
                     \midrule[0.08em]
                    baseline (CE) &59.6 &49.2\\
                     \hline
                    $\boldsymbol{\alpha} = \boldsymbol{0}$ &55.7  &47.1   \\
                    $\boldsymbol{\beta} = \boldsymbol{0}$ &63.2  &57.5  \\
                    $\boldsymbol{\alpha} + \boldsymbol{\beta} = \boldsymbol{1}$ &64.8  &59.1  \\
                    \hline
                    $\boldsymbol{\alpha}, \boldsymbol{\beta} \ge \boldsymbol{0}$ & \textbf{71.8}  &\textbf{67.3}  \\
                    \bottomrule[0.1em]
\end{tabular}
    }
    \vspace{-1.em}
\end{table}

\paragraph{The number of clean validation samples}
In Table~\ref{tab:validation-size}, our L2B method is shown to require few to no validation samples for LNL problems, highlighting its practicality. In the absence of a dedicated validation set, L2B adeptly generates an online meta set directly from the training data using a Gaussian mixture model~\cite{permuter2006study}, following~\cite{li2020dividemix}. L2B consistently boosts baseline methods such as DivideMix, UniCon, and C2D. Specifically, L2B-DivideMix has showcased its efficacy, particularly at high noise levels. Specifically, in a scenario with 90\% noise on CIFAR-10, our approach outstripped the baseline by 8.7\%, achieving an accuracy of 82.6\% compared to 76.0\%, and this was achieved without the need for clean validation samples. The advantage of L2B-DivideMix becomes even more pronounced when we incorporate a minimal amount of clean labels. With just 500 clean labels (equivalent to 2\% of the training data), our performance lead over the baseline extends to a remarkable 15.1\%. However, as we double the clean samples to 1000, the incremental benefit tapers off, yielding a mere 0.2\% boost. This behavior underscores the efficiency of L2B-DivideMix, demonstrating that it can deliver impressive results with minimal or even no clean validation data, making it a highly adaptable and practical solution for real-world applications.

\section{Conclusion}

Our paper presents Learning to Bootstrap (L2B), a new technique using joint reweighting for model training. L2B dynamically balances weights between actual labels, pseudo-labels, and different samples, mitigating the challenges of erroneous pseudo-labels.  Notably, L2B operates effectively without a clean validation set and can be well generalized to other tasks, highlighting its practicality in real-world settings. Extensive experiments on CIFAR-10, CIFAR-100, ISIC2019, and Clothing 1M datasets demonstrate the superiority and robustness compared to other existing methods under various settings.

\vspace{.5em}
\paragraph{Acknowledgments}
This work is partially supported by the AWS Public Sector Cloud Credit for Research Program.

\newpage

{
    \small
    \bibliographystyle{ieeenat_fullname}
    \bibliography{main}
}

\clearpage
\setcounter{page}{1}
\maketitlesupplementary

\section{Appendix}

\subsection{Normalization function comparision.}\label{A.1} 
\begin{figure}[h]

\centering
        \includegraphics[width=\linewidth]{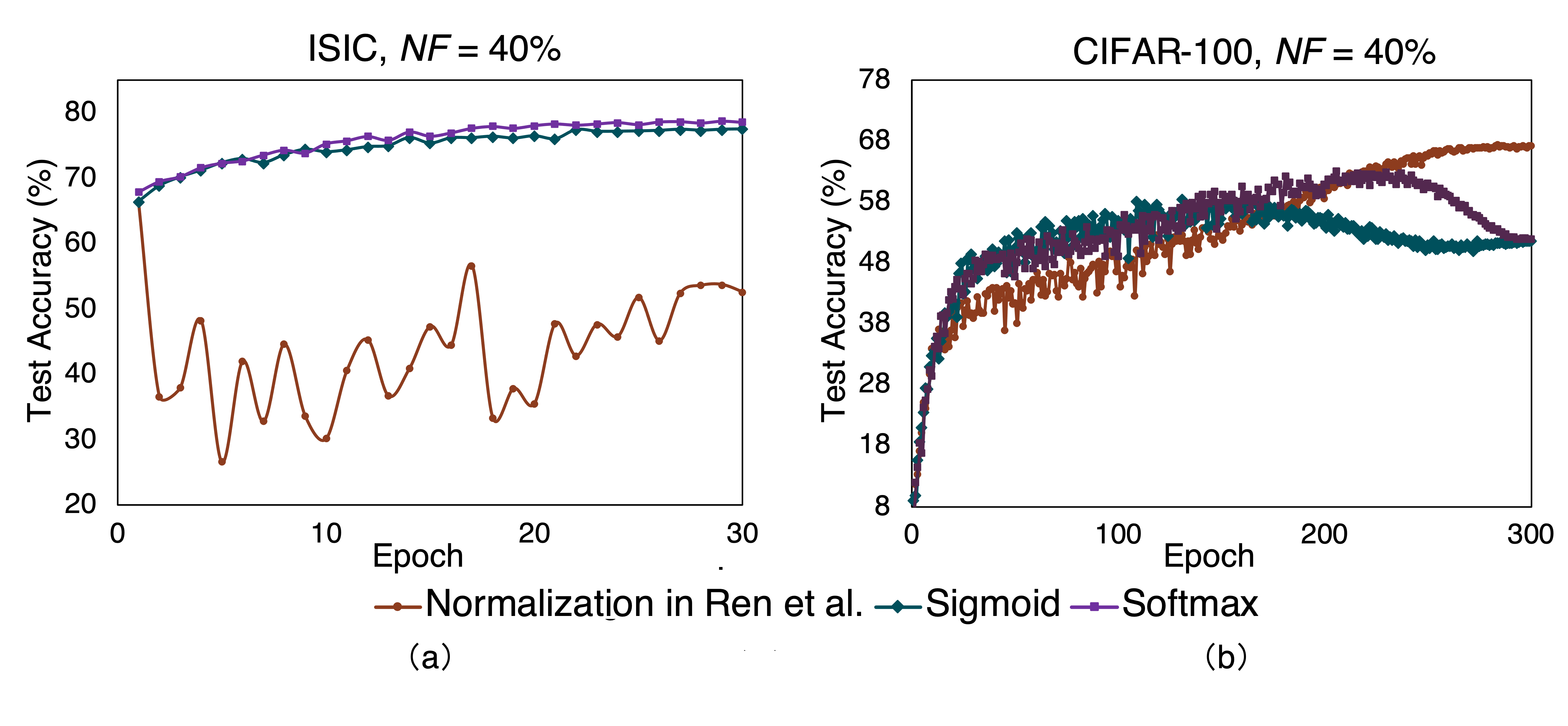}
        \vspace{-1.5em}
        \captionof{figure}{Comparison among different normalization functions (i.e., Eq.~\ref{eq:alpha_norm}, Sigmoid function and Softmax function). Testing accuracy curve: (a) with different normalization functions under 40\% symmetric noise label on the ISIC dataset. (b)  with different normalization under 40\% symmetric label noise on CIFAR-100.}
        \label{fig:normalization_comparison}
        \vspace{-1em}
\end{figure} 

\subsection{Alleviate potential overfitting to noisy examples.}\label{A.2} 
\begin{figure}[h]
\centering
\includegraphics[width=0.75\linewidth]{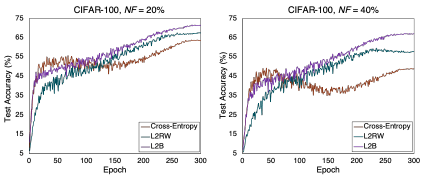}
\captionof{figure}{Test accuracy v.s. number of epochs on CIFAR-100 under the noise fraction of 20\%  and 40\%.}
\label{fig:accuracy_comparison}
\end{figure}

We also plot the testing accuracy curve under different noise fractions in Figure~\ref{fig:accuracy_comparison}, which shows that our proposed L2B would help preventing potential overfitting to noisy samples compared with standard training.
Meanwhile, compared to simply sample reweighting (L2RW), our L2B introduces pseudo-labels for bootstrapping the learner and is able to converge to a better optimum.

\begin{figure}[]
        \includegraphics[width=\linewidth]{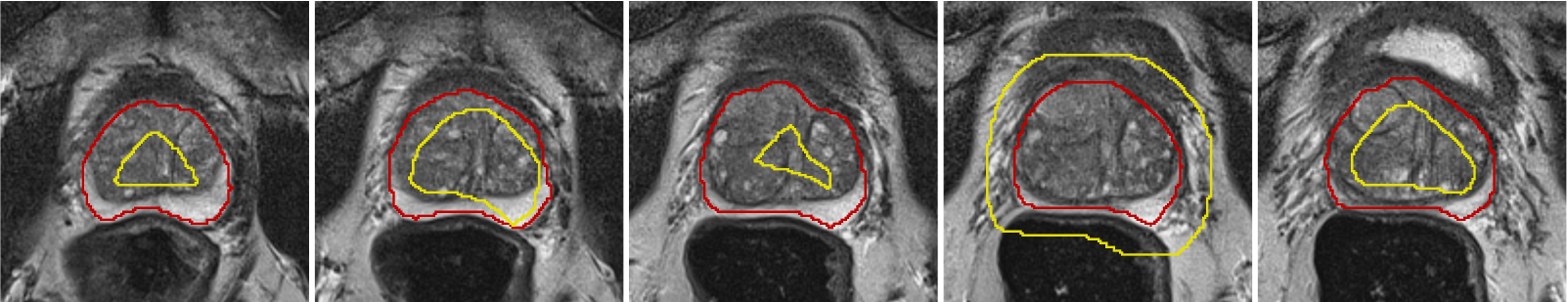}
         \caption{Visual comparison of prostate MRI images with noisy (contoured in yellow) and accurate (contoured in red) segmentation masks to demonstrate the discrepancy in segmentation quality between the two.}
        \label{fig:noisy_vis}
\end{figure}

\section{Theoretical Analysis}
\subsection{Equivalence of the two learning objectives}\label{B.1} 
We show that Eq.~\ref{eq:theta_star} is equivalent with Eq.~\ref{Eqn:reed_hard_loss} when $\forall i\  \alpha_i+\beta_i=1$. For convenience, we denote $y_i^{\text{real}},y_i^{\text{pseudo}},\mathcal{F}(x_i,\theta)$ using $y_i^r,y_i^p,p_i$ respectively. 
\begin{align}
&\alpha_i\mathcal{L}(p_i,y_i^r)+\beta_i\mathcal{L}(p_i,y_i^p)=\sum_{l=1}^L\alpha_i y_{i,l}^r\log p_{i,l}\\&+\beta_i y_{i,l}^p\log p_{i,l} 
=\sum_{l=1}^L(\alpha_i y_{i,l}^r+\beta_i y_{i,l}^p)\log p_{i,l}
\end{align}
Due to that $\mathcal{L}(\cdot)$ is the cross-entropy loss, we have $\sum_{l=1}^L y_{i,l}^r=\sum_{l=1}^L y_{i,l}^p=1$. Then $\sum_{l=1}^L\alpha_i y_{i,l}^r+\beta_i y_{i,l}^p=\alpha_i+\beta_i$. So if $\alpha_i+\beta_i=1$, we have 
\begin{align}
\sum_{l=1}^L(\alpha_i y_{i,l}^r+\beta_i y_{i,l}^p)&\log p_{i,l}=\mathcal{L}(p_i,\alpha_i y_i^r+\beta_i y_i^p)\\
&=\mathcal{L}(p_i,(1-\beta_i) y_i^r+\beta_i y_i^p)
\end{align}

\subsection{Gradient used for updating $\theta$}\label{B.2} 
We derivative the update rule for $\boldsymbol{\alpha},\boldsymbol{\beta}$ in Eq.~\ref{eq:theta_t+1}. 
\begin{align}
\alpha_{t,i}&=-\eta\frac{\partial}{\partial \alpha_i}(\sum_{j=1}^m f_j^v(\hat{\theta}_{t+1}))\Big |_{\alpha_i=0} \\
&=-\eta\sum_{j=1}^m \nabla f_j^v(\hat{\theta}_{t+1})^T\frac{\partial \hat{\theta}_{t+1}}{\partial \alpha_i}\Big |_{\alpha_i=0} \\ 
&=-\eta\sum_{j=1}^m \nabla f_j^v(\hat{\theta}_{t+1})^T\\
&\frac{\partial( \theta_t - \lambda \nabla (\sum_k \alpha_k \ f_{k}(\theta)+\beta_k\ g_{k}(\theta))\Big |_{\theta=\theta_t}) }{\partial \alpha_i}\Big |_{\alpha_i=0}\\
&=\eta\lambda\sum_{j=1}^m \nabla f_j^v(\theta_t)^T \nabla f_i(\theta_t)
\end{align}

\begin{align}
\beta_{t,i}&=-\eta\frac{\partial}{\partial \beta_i}(\sum_{j=1}^m f_j^v(\hat{\theta}_{t+1}))\Big |_{\beta_i=0} \\
&=-\eta\sum_{j=1}^m \nabla f_j^v(\hat{\theta}_{t+1})^T\frac{\partial \hat{\theta}_{t+1}}{\partial \beta_i}\Big |_{\beta_i=0} \\ 
&=-\eta\sum_{j=1}^m \nabla f_j^v(\hat{\theta}_{t+1})^T \\
&\frac{\partial( \theta_t - \lambda \nabla (\sum_k \alpha_k \ g_{k}(\theta)+\beta_k\ g_{k}(\theta))\Big |_{\theta=\theta_t}) }{\partial \beta_i}\Big |_{\beta_i=0}\\
&=\eta\lambda\sum_{j=1}^m \nabla f_j^v(\theta_t)^T \nabla g_i(\theta_t)
\end{align}

Then $\theta_{t+1}$ can be calculated by Eq.~\ref{eq:theta_t+1} using the updated $\alpha_{t,i},\beta_{t,i}$. 
\subsection{Convergence}\label{B.3} 
This section provides the proof for covergence (Section~\ref{sec:convergence}).

\begin{theorem*}
Suppose that the training loss function $f,g$ have $\sigma$-bounded gradients and the validation loss $f^v$ is Lipschitz smooth with constant L. With a small enough learning rate $\lambda$, the validation loss monotonically decreases for any training batch $B$, namely, 
\begin{equation}\label{eq:Validation_loss_decrease_app}
G(\theta_{t+1})\leq G(\theta_t),
\end{equation}
where $\theta_{t+1}$ is obtained using Eq.~\ref{eq:theta_t+1} and $G$ is the validation loss
\begin{equation}\label{eq:Validation_loss_app}
G(\theta)=\frac{1}{M}\sum_{i=1}^M f_i^v(\theta),
\end{equation}

Furthermore, Eq.~\ref{eq:Validation_loss_decrease_app} holds for all possible training batches only when the gradient of validation loss function becomes $0$ at some step $t$, namely, $G(\theta_{t+1})=G(\theta_t)\ \forall B \Leftrightarrow \nabla G(\theta_t)=0$
\end{theorem*}

\textit{Proof.} At each training step $t$, we pick a mini-batch $B$ from the union of training and validation data with $|B|=n$. From section B we can derivative $\theta_{t+1}$ as follows: 
\begin{align}
\theta_{t+1}&=\theta_t-\lambda\sum_{i=1}^n(\alpha_{t,i}\nabla f_i(\theta_t)+\beta_{t,i}\nabla g_i(\theta_t)) \\
&=\theta_t-\eta\lambda^2M\sum_{i=1}^n(\nabla G^T\nabla f_i \nabla f_i+\nabla G^T\nabla g_i \nabla g_i)
\end{align}

We omit $\theta_t$ after every function for briefness and set $m$ in section B equals to $M$. Since $G(\theta)$ is Lipschitz-smooth, we have 
\begin{equation}
    G(\theta_{t+1})\leq G(\theta_t)+\nabla G^T \Delta \theta+\frac{L}{2}||\Delta \theta||^2.
\end{equation}

Then we show $\nabla G^T \Delta \theta+\frac{L}{2}||\Delta \theta||^2\leq 0$ with a small enough $\lambda$. Specifically, \begin{equation}
\nabla G^T \Delta \theta = -\eta\lambda^2M\sum_i(\nabla G^T \nabla f_i)^2+(\nabla G^T \nabla g_i)^2.
\end{equation}

Then since $f_i,g_i$ have $\sigma$-bounded gradients, we have 
\begin{align}
\frac{L}{2}||\Delta \theta||^2&\leq \frac{L\eta^2\lambda^4M^2}{2}\sum_i (\nabla G^T \nabla f_i)^2||\nabla f_i||^2 \\
&+(\nabla G^T \nabla g_i)^2||\nabla g_i||^2\\
&\leq \frac{L\eta^2\lambda^4M^2\sigma^2}{2}\sum_i (\nabla G^T \nabla f_i)^2 +(\nabla G^T \nabla g_i)^2
\end{align}

Then if $\lambda^2 < \frac{2}{\eta \sigma^2 ML}$, 
\begin{align}
\nabla G^T \Delta \theta+\frac{L}{2}||\Delta \theta||^2&\leq (\frac{L\eta^2\lambda^4M^2\sigma^2}{2}-\eta\lambda^2M) \\&
\sum_i(\nabla G^T \nabla f_i)^2+(\nabla G^T \nabla g_i)^2\leq 0.
\end{align}

Finally we prove $G(\theta_{t+1})=G(\theta_t)\ \forall B \Leftrightarrow \nabla G(\theta_t)=0$: 
If $\nabla G(\theta_t)=0$, from section B we have $\alpha_{t,i}=\beta_{t,i}=0$, then $\theta_{t+1}=\theta_t$ and thus $G(\theta_{t+1})=G(\theta_t)\ \forall B$. Otherwise, if $\nabla G(\theta_t)\neq 0$, we have 
\begin{equation}
    0<||\nabla G||^2=\nabla G^T \nabla G = \frac{1}{M}\sum_{i=1}^M\nabla G^T \nabla f_i^v,
\end{equation}
which means there exists a $k$ such that $\nabla G^T\nabla f_k^v>0$. So for the mini-batch $B_k$ that contains this example, we have

\begin{align}
    G(\theta_{t+1})-G(\theta_t)&\leq \nabla G^T\Delta\theta+\frac{L}{2}||\Delta\theta||^2 \\
    &\leq (\frac{L\eta^2\lambda^4M^2\sigma^2}{2}-\eta\lambda^2M)\\
    &\sum_{i\in B}(\nabla G^T\nabla f_i)^2+(\nabla G^T\nabla g_i)^2 \\
    &\leq (\frac{L\eta^2\lambda^4M^2\sigma^2}{2}-\eta\lambda^2M) \nabla G^T \nabla f_k^v \\
    &<0.
\end{align}

\end{document}